\pdfoutput=1

\documentclass[11pt]{article}

\usepackage[preprint]{acl}

\usepackage{times}
\usepackage{latexsym}

\usepackage[T1]{fontenc}

\usepackage[utf8]{inputenc}

\usepackage{microtype}

\usepackage{inconsolata}

\usepackage{graphicx}

\usepackage{subcaption}
\usepackage{booktabs}
\usepackage{multirow}
\usepackage{amsmath}
\usepackage{amsfonts}
\usepackage{bm}
\usepackage{makecell}
\usepackage{threeparttable}
\usepackage{cleveref}

\newcommand{\subfigwidth}{0.32\textwidth}

\crefname{equation}{Equation}{Equations}
\crefname{figure}{Figure}{Figures}
\crefname{table}{Table}{Tables}
\crefname{section}{\S}{\S}
\crefname{subfigure}{Figure}{Figures}

\title{Instability in Downstream Task Performance During LLM Pretraining}

\author{
  Yuto Nishida$^{1,3}$ $\;\;\;$ 
  Masaru Isonuma$^{3,2}$ $\;\;\;$
  Yusuke Oda$^{1,3}$ $\;\;\;$ 
  \\
  $^1$Nara Institute of Science and Technology $\;$ 
  $^2$Tohoku University$\;$ \\
  $^3$Research and Development Center for Large Language Models, \\ National Institute of Informatics$\;$ \\
  \texttt{\{nishida.yuto.nu8, yusuke.oda\}@naist.ac.jp} \\
  \texttt{isonuma@nii.ac.jp}
}

\begin{document}
\maketitle
\begin{abstract}
When training large language models (LLMs), it is common practice to track downstream task performance throughout the training process and select the checkpoint with the highest validation score.
However, downstream metrics often exhibit substantial fluctuations, making it difficult to identify the checkpoint that truly represents the best-performing model.
In this study, we empirically analyze the stability of downstream task performance in an LLM trained on diverse web-scale corpora.
We find that task scores frequently fluctuate throughout training, both at the aggregate and example levels.
To address this instability, we investigate two post-hoc checkpoint integration methods: checkpoint averaging and ensemble, motivated by the hypothesis that aggregating neighboring checkpoints can reduce performance volatility.
We demonstrate both empirically and theoretically that these methods improve downstream performance stability without requiring any changes to the training procedure.
\end{abstract}

\section{Introduction}
Large language models (LLMs) are typically developed through pretraining on large-scale corpora, during which the performance on downstream tasks evolves dynamically.
Understanding these training dynamics is essential for diagnosing issues during model development and ensuring the reliability, reproducibility, and effectiveness of the resulting models on downstream tasks.
In particular, the stability of evaluation metrics plays a key role in model development: when evaluation metrics fluctuate erratically, they undermine the reliability of the evaluation and hamper fair comparison across checkpoints, models, or experimental settings.
Such instability may arise not only in evaluation signals based on next-token prediction, such as training loss, but also in downstream task metrics that are central to model evaluation.

\begin{figure}[t]
\includegraphics[width=1.0\linewidth, keepaspectratio]{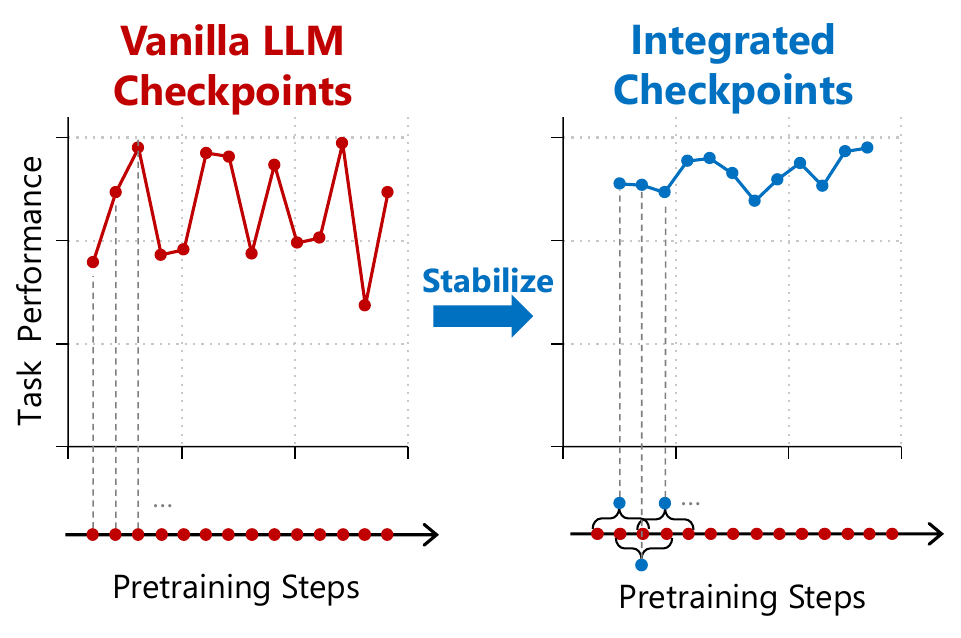}
\caption{Overview of stabilizing downstream task performance during pretraining. We observe instability in task performance during LLM pretraining, and experimentally show that theoretically motivated checkpoint integration methods improve evaluation stability.}
\label{fig:overview}
\end{figure}

Previous research on training stability has largely focused on signals derived from the next-token prediction objective.
Studies of loss dynamics have offered insights into how architectural design, optimization strategies, and initialization schemes can encourage smooth convergence~\cite{chowdhery2023palm,rybakov2024methodsimprovingllmtraining,takase2024spikemorestabilizingpretraining}.
More recently, \citet{chang-etal-2024-characterizing} have investigated token-level prediction probabilities and shown that their stability varies with token frequency and contextual diversity.

Practical evaluation of LLMs typically relies on downstream tasks such as generation, question answering, or classification, while prior studies shed light on dynamics related to the next-token prediction objective.
For reliable evaluation, it is essential to understand how stable downstream performance is during pretraining and to explore ways to mitigate instability when it arises.
However, little is known about how downstream task performance behaves throughout the pretraining process.

To this end, we empirically analyze the stability of LLMs of different sizes pretrained on a diverse web-scale corpus.  
By tracking the performance of multiple downstream tasks across a series of training checkpoints, we observe that scores generally improve in the long run as training progresses, while in the short term they repeatedly rise and fall in an inconsistent manner, which we refer to as \emph{fluctuations}.  
These fluctuations are observed not only in aggregate task-level scores but also at the level of individual examples, indicating that instability is inherent in downstream performance during LLM pretraining.  
We further examine the effect of model size on this instability and find that scaling up models does not necessarily mitigate fluctuations.  
This inherent instability, in turn, hinders robust and reliable evaluation.

To address this challenge, we explore methods for mitigating such fluctuations in downstream performance.
Motivated by the hypothesis that aggregating multiple checkpoints can smooth out performance volatility during training, we investigate two checkpoint integration methods: checkpoint averaging and ensemble.
These methods, while simple and theoretically motivated, can reduce short-term instability without altering the training process.  
Our experimental results show that these methods can mitigate performance instability and improve evaluation robustness during training.

\begin{figure*}[t]
\captionsetup[subfigure]{justification=centering, skip=3pt}
  \centering
  
  \begin{minipage}[b]{\subfigwidth}
    \centering
    \includegraphics[width=\textwidth]{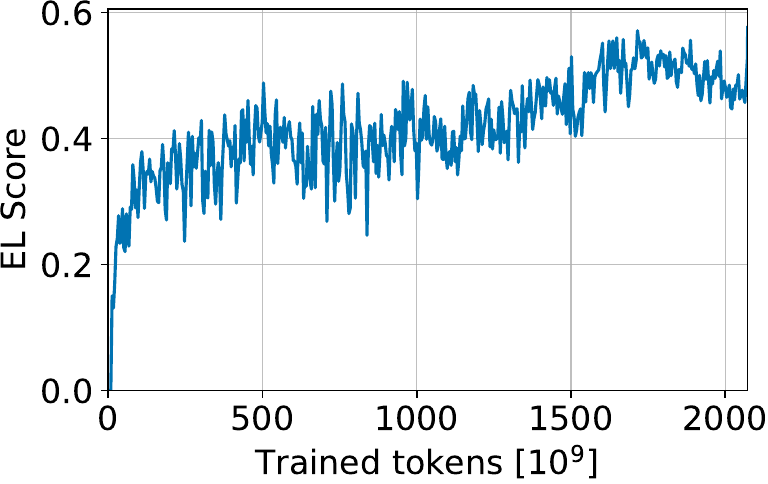}
    \subcaption{EL (Entity Linking)}\label{fig:el}
  \end{minipage}
  \hfill
  \begin{minipage}[b]{\subfigwidth}
    \centering
    \includegraphics[width=\textwidth]{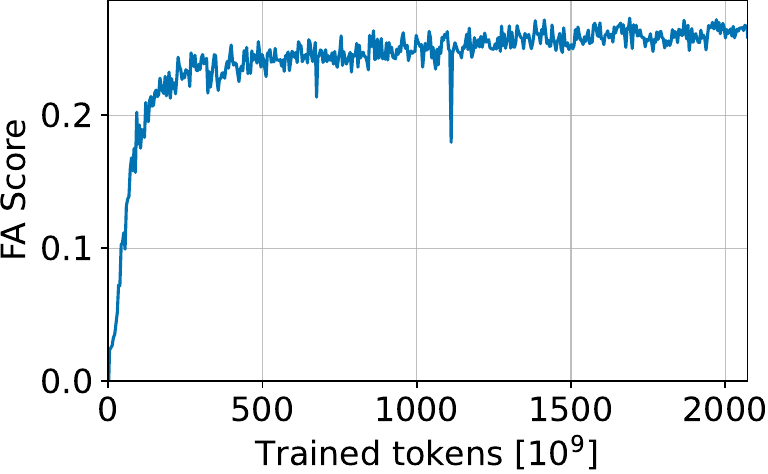}
    \subcaption{FA (Fundamental Analysis)}\label{fig:fa}
  \end{minipage}
  \hfill
  \begin{minipage}[b]{\subfigwidth}
    \centering
    \includegraphics[width=\textwidth]{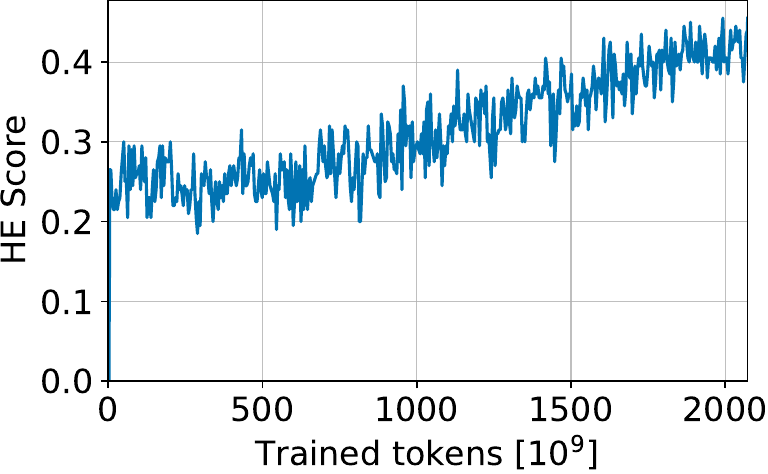}
    \subcaption{HE (Human Examination)}\label{fig:he}
  \end{minipage}

  \vspace{1em}

  \begin{minipage}[b]{\subfigwidth}
    \centering
    \includegraphics[width=\textwidth]{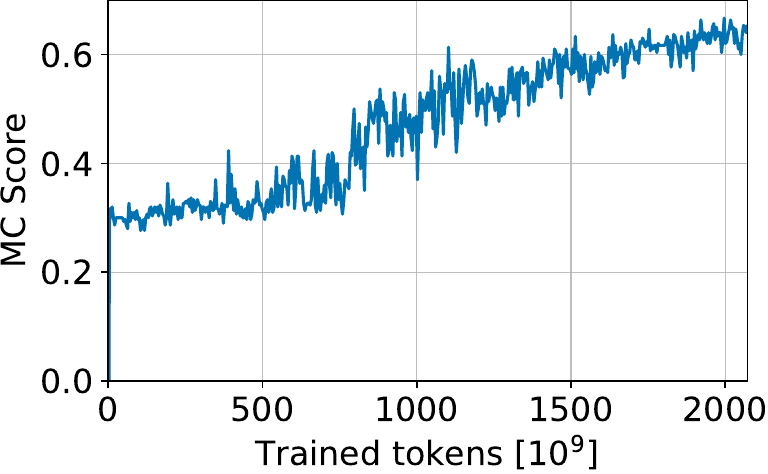}
    \subcaption{MC (Multiple Choice QA)}\label{fig:mc}
  \end{minipage}
  \hfill
  \begin{minipage}[b]{\subfigwidth}
    \centering
    \includegraphics[width=\textwidth]{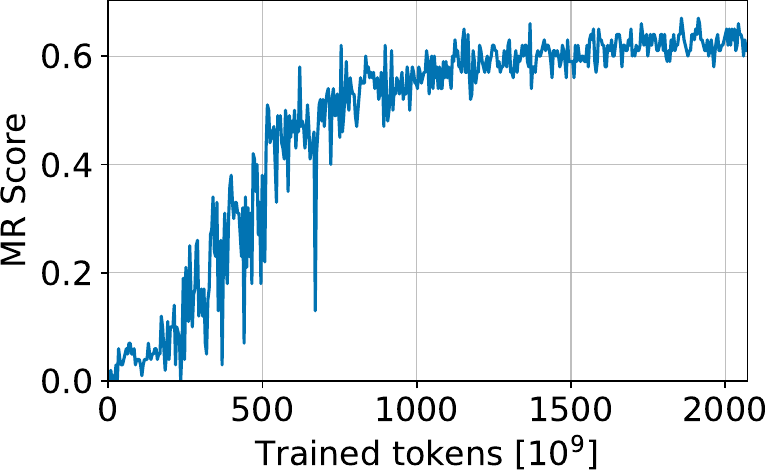}
    \subcaption{MR (Mathematical Reasoning)}\label{fig:mr}
  \end{minipage}
  \hfill
  \begin{minipage}[b]{\subfigwidth}
    \centering
    \includegraphics[width=\textwidth]{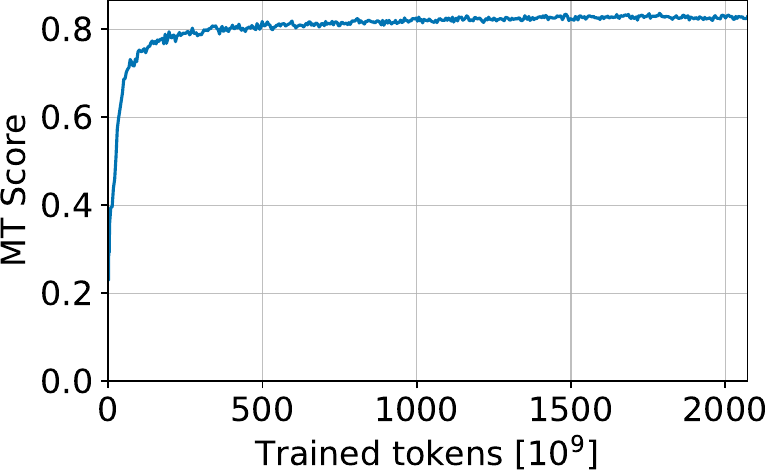}
    \subcaption{MT (Machine Translation)}\label{fig:mt}
  \end{minipage}

  \vspace{1em}

  \begin{minipage}[b]{\subfigwidth}
    \centering
    \includegraphics[width=\textwidth]{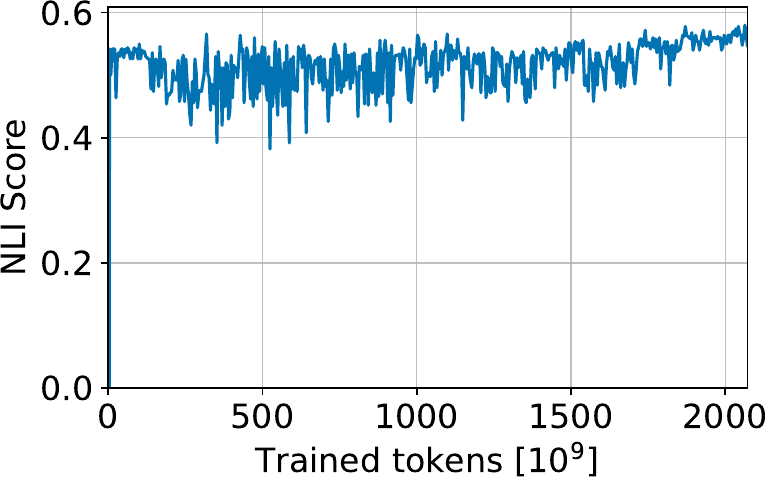}
    \subcaption{NLI (Natural Language Inference)}\label{fig:nli}
  \end{minipage}
  \hfill
  \begin{minipage}[b]{\subfigwidth}
    \centering
    \includegraphics[width=\textwidth]{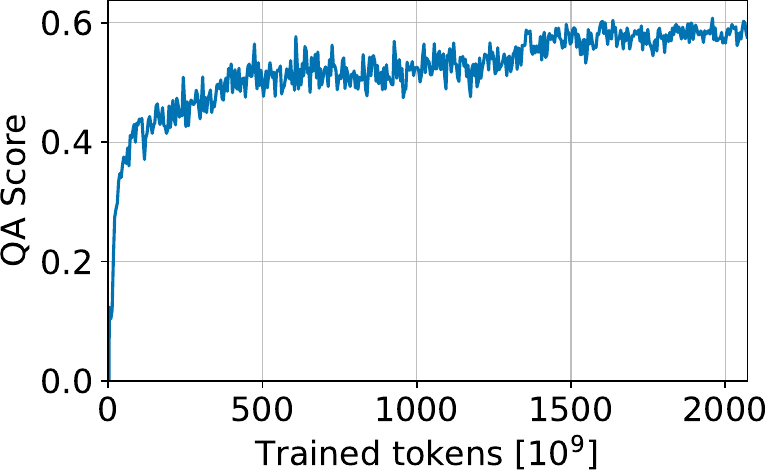}
    \subcaption{QA (Question Answering)}\label{fig:qa}
  \end{minipage}
  \hfill
  \begin{minipage}[b]{\subfigwidth}
    \centering
    \includegraphics[width=\textwidth]{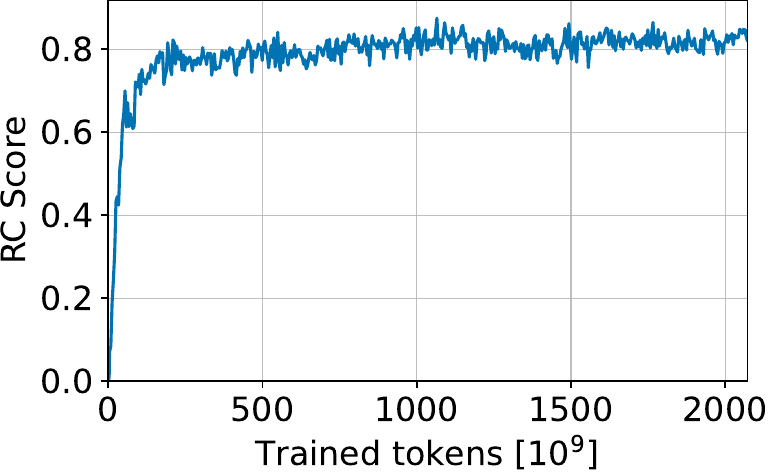}
    \subcaption{RC (Reading Comprehension)}\label{fig:rc}
  \end{minipage}

\caption{Score trajectories of the 13B model across downstream task categories. In most categories, we observe gradual long-term improvements over the course of training, accompanied by short-term score fluctuations that persist throughout.}
\label{fig:all_categories}
\end{figure*}

\begin{figure}[t]
\centering
\includegraphics[width=0.9\linewidth, keepaspectratio]{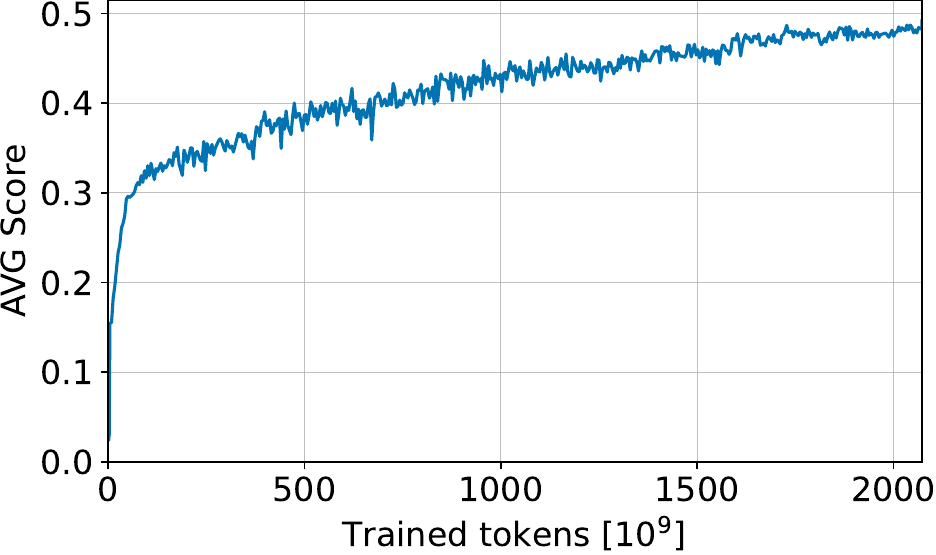}
\caption{Average score trajectory of the 13B model across all downstream task categories.}
\label{fig:average_category}
\end{figure}

\section{Observation of Task Instability}
\label{sec:phenomena}

\subsection{Experimental Settings}
We conduct our analysis using a family of Transformer-based language models trained on web-scale corpora containing a mixture of Japanese, English, and source code.  
The training data, totaling approximately 2.1 trillion tokens after weighted sampling, is drawn from the LLM-jp Corpus v3.\footnote{\url{https://gitlab.llm-jp.nii.ac.jp/datasets/llm-jp-corpus-v3}}  
We analyze seven pre-trained models with varying the number of parameters: 150M, 440M, 980M, 1.8B, 3.7B, 7.2B, and 13B.  
During pretraining, checkpoints were regularly obtained at intervals ranging from several hundred to approximately a thousand steps for each model, according to predefined schedules detailed in Appendix~\ref{appendix:detailed_settings}.
In our analysis, we use all checkpoints except the initial one (step 0), where parameters are randomly initialized.\footnote{At the first step, the learning rate is set to zero, so parameters at step 0 and step 1 are identical; the only difference is that the optimizer state is updated.}
Our analysis focuses on the pretraining phase, and does not include instruction-tuned or otherwise adapted models.

For evaluation, we use the llm-jp-eval v1.4.1\footnote{\url{https://github.com/llm-jp/llm-jp-eval}} to perform 4-shot inference on a suite of downstream tasks.
We evaluate nine task categories: EL (entity linking), FA (fundamental analysis), HE (human examination), MC (multiple choice question answering), MR (mathematical reasoning), MT (machine translation), NLI (natural language inference), QA (question answering), and RC (reading comprehension).
Each category typically includes multiple datasets; for example, the MC category includes three datasets.
Details of the tasks, datasets, and evaluation metrics used in each category are provided in Appendix~\ref{appendix:detailed_settings}.

\subsection{Instability Across Task Categories}
We begin by examining downstream performance across task categories over the course of pretraining.
\cref{fig:all_categories} shows score trajectories for the 13B model across task categories.
In all categories except for machine translation (MT; shown in \cref{fig:mt}), we observe persistent short-term score fluctuations throughout training.\footnote{In the following sections, we focus on the eight categories other than MT, where instability was qualitatively observed.}
Similar fluctuations are also observed in \cref{fig:average_category}, which presents the trajectory of average scores across all categories.
These fluctuations are not limited to a particular phase of training, such as early instability or late saturation, but instead persist across many checkpoints.
Similar trends are observed across models of other sizes, indicating that instability in task performance is a general phenomenon not limited to specific scales.

\subsection{Example-Level Score Dynamics}
\begin{figure*}[t]
\captionsetup[subfigure]{justification=centering}
  \centering
  \begin{minipage}[b]{\linewidth}
    \centering
    \includegraphics[width=1.0\linewidth]{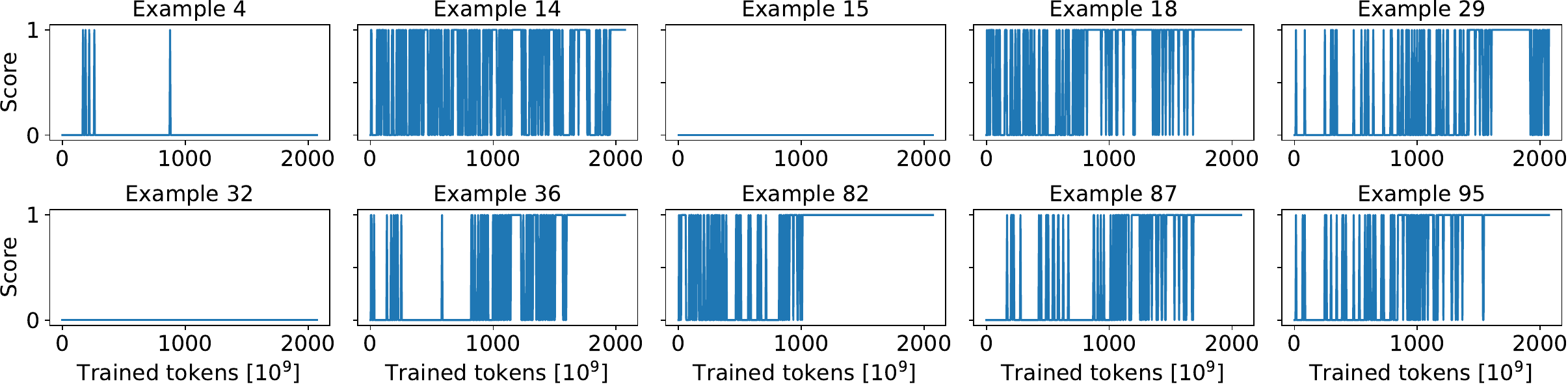}
    \subcaption{MC: JCommonsenseQA}\label{fig:13b_jcommonsenseqa_example_level}
  \end{minipage}
  \vspace{0.8em}
  \begin{minipage}[b]{\linewidth}
    \centering
    \includegraphics[width=1.0\linewidth]{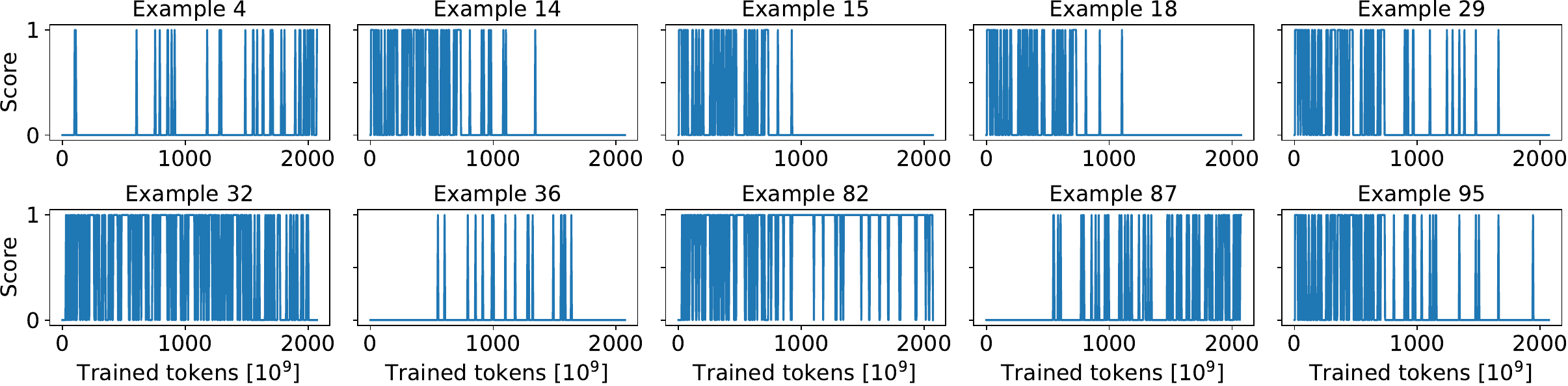}
    \subcaption{NLI: Jamp}\label{fig:13b_jamp_example_level}
  \end{minipage}
  \vspace{0.8em}
  \begin{minipage}[b]{\linewidth}
    \centering
    \includegraphics[width=1.0\linewidth]{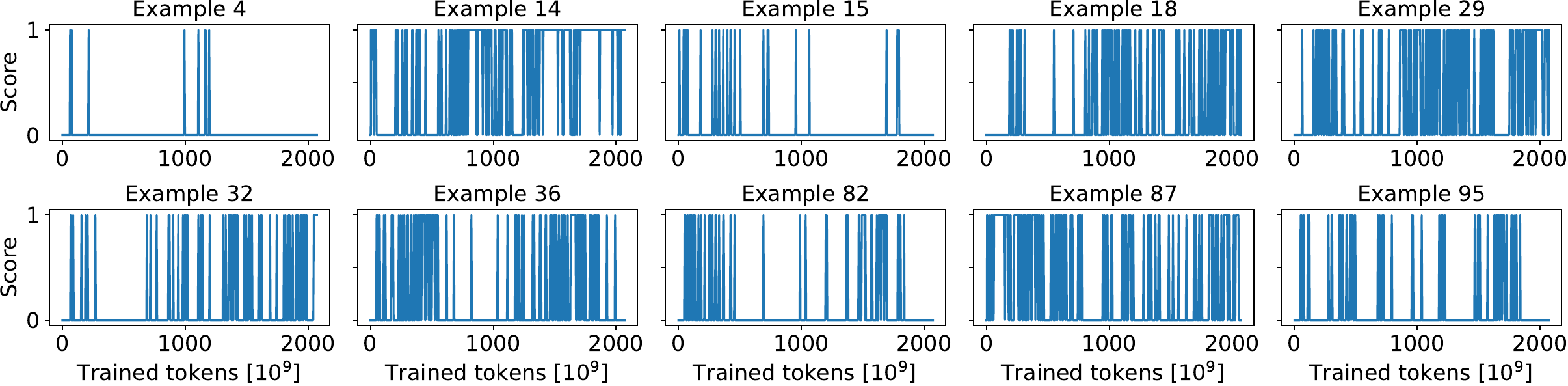}
    \subcaption{HE: JMMLU}\label{fig:13b_jmmlu_example_level}
  \end{minipage}
\caption{Example-level score trajectories of the 13B model during training. Across examples from MC, NLI, and HE categories, we observe frequent alternations between correct and incorrect predictions, indicating persistent prediction instability over time.}
\label{fig:example_level_score}
\end{figure*}

To understand the source of task-level fluctuations, we investigate example-level prediction dynamics.
\cref{fig:example_level_score} shows score trajectories for ten examples drawn from JCommonsenseQA~\cite{kurihara-etal-2022-jglue} in the MC category, Jamp~\cite{sugimoto-etal-2023-jamp} in the NLI category, and JMMLU~\cite{yin-etal-2024-respect} in the HE category using the 13B model.\footnote{We uniformly sampled indices, and then plotted the score trajectories of the examples corresponding to those indices.}
Since these tasks use exact match as the evaluation metric, the score for each example is either 0 (incorrect) or 1 (correct). Across many examples, we observe frequent alternations between correct and incorrect predictions, even over extended training periods.
While some examples converge to stable predictions, many continue to exhibit prediction instability well beyond early training.

These example-level dynamics suggest that the task-level score fluctuations observed earlier may be attributed to the accumulation of such local prediction instabilities across many examples.

\subsection{Scoring-Based Analysis of Pretraining Instability}
\label{sec:instability_quantification}
\begin{figure*}[t]
\centering
\begin{minipage}[b]{0.48\linewidth}
  \centering
  \includegraphics[width=\linewidth]{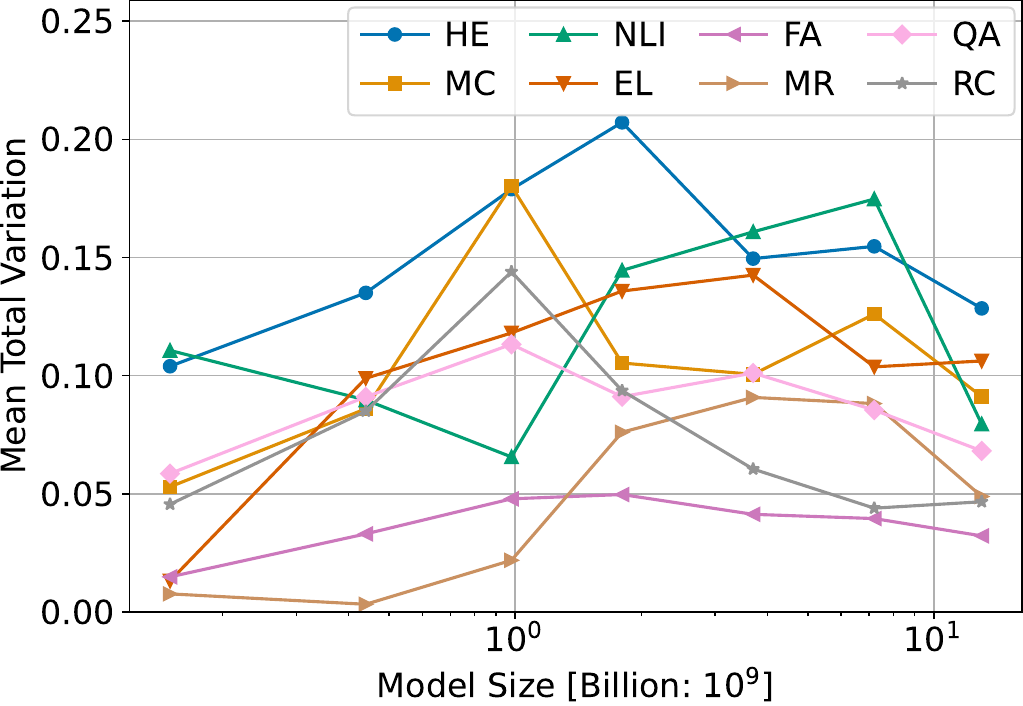}
  \subcaption{Mean total variation (MTV)}\label{fig:tv_by_model_size}
\end{minipage}
\hfill
\begin{minipage}[b]{0.48\linewidth}
  \centering
  \includegraphics[width=\linewidth]{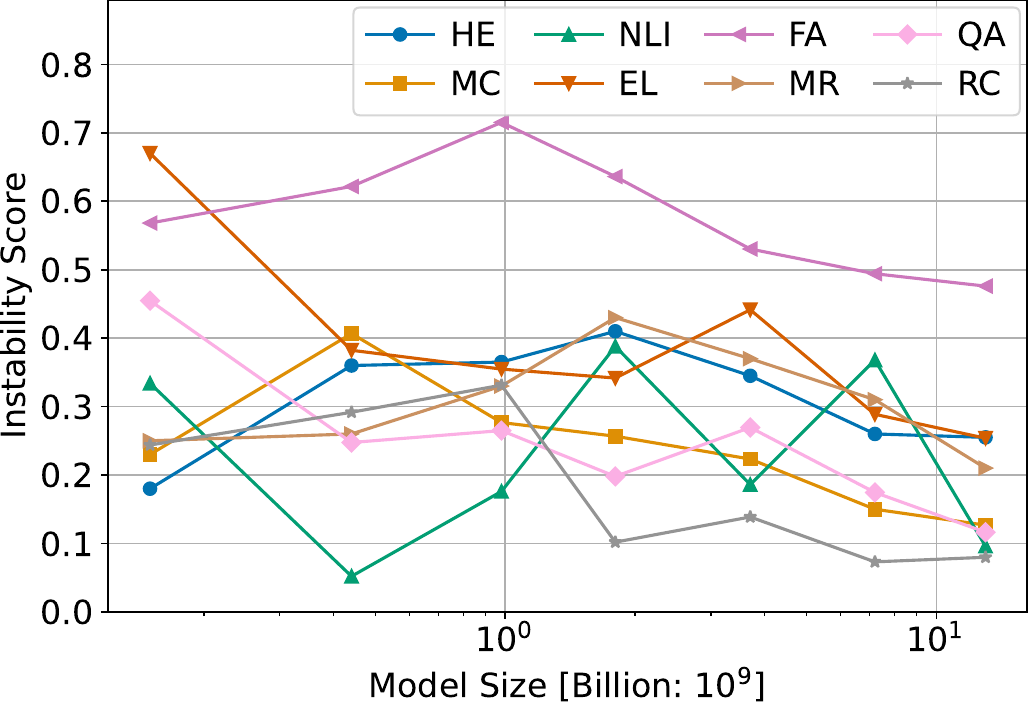}
  \subcaption{Instability score (IS)}\label{fig:instability_score_by_model_size}
\end{minipage}
\caption{Relationship between model size and two metrics of downstream performance instability. MTV is computed from reference-based evaluation scores, while IS captures output dissimilarity between consecutive checkpoints. Neither metric shows a consistent decreasing trend with increasing model size, suggesting that scaling does not reliably mitigate instability.}
\label{fig:instability_by_model_size}
\end{figure*}

To quantify instability of downstream task performance across checkpoints, we begin by measuring fluctuations in reference-based evaluation scores.
Let $\bm{\Theta} = \{\bm{\theta}_1, \ldots, \bm{\theta}_m\}$ denote the sequence of $m$ checkpoints obtained during pretraining.  
Let $x_{\bm{\theta}_t}$ be the output generated using parameters $\bm{\theta}_t$ for a given input, and let $f(x_{\bm{\theta}_t})$ denote the reference-based evaluation score for that output (e.g., exact match or charF1).  
We denote $L(\bm{\theta}_t) = \mathbb{E}_{x_{\bm{\theta}_t} \sim p(x;\bm{\theta}_t)}[f(x_{\bm{\theta}_t})]$ as the expected value of the evaluation score given an output sampled from the model $\bm{\theta}_t$.
We use total variation over these scores to measure how predictions change throughout training.  
Specifically, we compute the mean total variation (MTV) as follows:
\begin{equation}
\mathrm{MTV}(\bm{\Theta}) = \frac{1}{m - 1} \sum_{t=1}^{m - 1} \left| L(\bm{\theta}_{t+1}) - L(\bm{\theta}_t) \right|.
\end{equation}
This value captures the average magnitude of score transitions and is normalized by the number of transitions.
This normalization ensures fair comparison across models with different numbers of saved checkpoints.
In our experiments, we compute MTV using only the last 20\% of checkpoints, in order to focus on fluctuations that remain in the final stages of training, where model performance is typically evaluated and selected in practical settings.
\cref{fig:tv_by_model_size} shows the average MTV over the final training checkpoints for three representative task categories: MC, NLI, and HE.  
The results reveal no consistent relationship between model size and MTV, suggesting that scaling up model size does not reliably reduce instability in downstream performance.

Although MTV is useful for capturing reference-based score variability, it depends on specific evaluation metrics and does not directly assess how the model’s raw outputs change over time.
To address this, we introduce a complementary metric based on output similarity.  
For each example, we compute the dissimilarity between consecutive outputs $x_{\bm{\theta}_i}$ and $x_{\bm{\theta}_{i+1}}$ using a similarity function.\footnote{In this paper, we implemented $\mathrm{sim}(\cdot,\cdot)$ as a task-specific score function provided by the llm-jp-eval framework, such as exact match or charF1.}
We define the instability score (IS) as:
\begin{equation}
\mathrm{IS}(\bm{\Theta}) = \frac{1}{m - 1} \sum_{i=1}^{m - 1} \left( 1 - \mathrm{sim}(x_{\bm{\theta}_i}, x_{\bm{\theta}_{i+1}}) \right).
\end{equation}
In contrast to MTV, which reflects correctness with respect to references, this metric directly captures variability in the model's own output behavior, offering a complementary perspective on instability.  
We compute the IS using the same final 20\% of checkpoints as used for MTV, in order to evaluate the remaining output fluctuations at the end of training.
\cref{fig:instability_score_by_model_size} presents the average IS by model size. 
As with the MTV results, IS was not reduced by scaling up model size.

\begin{table*}[t]
\centering
\small
\tabcolsep 3pt
\begin{tabular}{@{}clccccccccc@{}}
\toprule
 & & \multicolumn{9}{c}{Task category} \\
\cmidrule(l){3-11}
\makecell{Stabilization\\method} & \makecell{Win.\\size} & Overall & EL & FA & HE & MC & MR & NLI & QA & RC \\
\midrule
NA & & $.477_{\pm .006}$ & $.403_{\pm .025}$ & $.621_{\pm .022}$ & $.547_{\pm .021}$ & $.507_{\pm .029}$ & $.261_{\pm .005}$ & $.625_{\pm .018}$ & $.579_{\pm .012}$ & $.820_{\pm .017}$ \\
\midrule
\multirow{5}{*}{Averaging}
& 2 & $.479_{\pm .005}$ & $.410_{\pm .022}$ & $.625_{\pm .018}$ & $.549_{\pm .020}$ & $.513_{\pm .026}$ & $.263_{\pm .005}$ & $.628_{\pm .019}$ & $.582_{\pm .011}$ & $.820_{\pm .017}$ \\
& 3 & $.480_{\pm .005}$ & $.412_{\pm .022}$ & $.628_{\pm .017}$ & $.551_{\pm .021}$ & $.512_{\pm .025}$ & $.263_{\pm .004}$ & $.628_{\pm .018}$ & $.584_{\pm .011}$ & $.819_{\pm .016}$ \\
& 5 & $.481_{\pm .005}$ & $.415_{\pm .022}$ & $.631_{\pm .016}$ & $.551_{\pm .019}$ & $.512_{\pm .023}$ & $.263_{\pm .004}$ & $.628_{\pm .016}$ & $.586_{\pm .010}$ & $.821_{\pm .014}$ \\
& 10 & $.483_{\pm .004}$ & $.423_{\pm .018}$ & $.632_{\pm .016}$ & $.553_{\pm .016}$ & $.518_{\pm .025}$ & $.264_{\pm .004}$ & $.628_{\pm .016}$ & $.592_{\pm .011}$ & $.819_{\pm .014}$ \\
& 20 & $.484_{\pm .003}$ & $.424_{\pm .013}$ & $.633_{\pm .013}$ & $.551_{\pm .015}$ & $.521_{\pm .020}$ & $.264_{\pm .003}$ & $.630_{\pm .010}$ & $.595_{\pm .009}$ & $.820_{\pm .010}$ \\
\midrule
\multirow{5}{*}{Ensemble}
& 2 & - & $.403_{\pm .021}$ & $.620_{\pm .018}$ & $.547_{\pm .019}$ & - & - & - & - & - \\
& 3 & - & $.408_{\pm .021}$ & $.626_{\pm .018}$ & $.549_{\pm .017}$ & - & - & - & - & - \\
& 5 & - & $.410_{\pm .020}$ & $.628_{\pm .015}$ & $.549_{\pm .018}$ & - & - & - & - & - \\
& 10 & - & $.414_{\pm .017}$ & $.631_{\pm .013}$ & $.548_{\pm .016}$ & - & - & - & - & - \\
& 20 & - & $.410_{\pm .012}$ & $.631_{\pm .015}$ & $.547_{\pm .015}$ & - & - & - & - & - \\
\bottomrule
\end{tabular}
\caption{Effect of stabilization methods on downstream task performance during the training of the 13B model. "Overall" indicates the aggregated score across all task categories. Scores are reported as mean ± standard deviation, computed over the final 20\% of training checkpoints. Both checkpoint averaging and ensemble reduce score variance and improve mean performance across many task categories.}

\label{tab:stabilized_score}
\end{table*}

\section{Mitigation of Instability Through Post-Processing}
In the previous section, we observed that downstream task performance often fluctuates during the pretraining of large language models (LLMs), both at the task and example levels. 
These fluctuations hinder reliable evaluation and model selection, especially when performance at the final training steps is used to benchmark models. 
To address this challenge, we investigate post-processing methods that stabilize performance by integrating information from multiple adjacent checkpoints. 
In the following, we describe two techniques, checkpoint averaging and checkpoint ensemble, provide a theoretical motivation for their use, and empirically evaluate their effectiveness in mitigating instability at both the task and example levels.

\subsection{Post-Processing Methods}
\paragraph{Checkpoint averaging} is a technique in which model weights from multiple adjacent checkpoints are averaged to produce a single inference-time model. 
At each training step $t$, we compute the average of the model parameters saved in the most recent $n$ checkpoints: $\bar{\bm{\theta}}_t = \frac{1}{m} \sum_{i=t-m+1}^{t} \bm{\theta}_i$.
Prior studies have shown that this approach can improve model stability and generalization~\cite{gao-etal-2022-revisiting}. In our experiments, we use window sizes $m$ of 2, 3, 5, 10, and 20.

\paragraph{Checkpoint ensemble} is an alternative that retains the original models and aggregates their inference outputs~\cite{chen2017checkpointensemblesensemblemethods,Liu2018ACS}.
Specifically, we use a majority vote ensemble, where the final prediction is determined by the most frequently predicted label among the constituent checkpoints.
Each checkpoint contributes equally.
This method is only applicable to multiple-choice tasks (MC, NLI, HE), where the output space consists of a small number of predefined options.
We use the same window sizes ($m = 2, 3, 5, 10, 20$) as in averaging.
Unlike averaging, checkpoint ensemble does not modify model parameters and is free from artifacts introduced by weight interpolation.
However, it incurs a higher inference cost, as it requires separate forward passes for each checkpoint.

\subsection{Theoretical Motivation}
We now provide a theoretical justification for why checkpoint averaging and ensemble would enhance the stability of downstream performance during training.  
Let $ \bm{\theta}_i $ denote the model parameters at training step $ i $, and define the average over the most recent $n$ checkpoints as $ \bar{\bm{\theta}}_t = \frac{1}{n} \sum_{i=t-n+1}^{t} \bm{\theta}_i $.  
We define $ \bar{\bm{\Theta}} = \{\bar{\bm{\theta}}_1, \ldots, \bar{\bm{\theta}}_m\} $ and show $ \mathrm{MTV}(\bar{\bm{\Theta}})\le \mathrm{MTV}(\bm{\Theta}) $.

Suppose each $ \bm{\theta}_i $ is close to $ \bar{\bm{\theta}}_t $, we can assume a first-order approximation of the evaluation score\footnote{Particularly in the final stage of pretraining, which is important in practice, changes in model parameters between adjacent checkpoints are sufficiently small, making this assumption reasonable.}:
\begin{align}
L(\bar{\bm{\theta}}_t) &\approx L(\bm{\theta}_i) + (\bar{\bm{\theta}}_t - \bm{\theta}_i)^\top \nabla_{\bm{\theta}} L(\bm{\theta}_i), \label{approx0} \\
L(\bm{\theta}_i) &\approx L(\bar{\bm{\theta}}_t) + (\bm{\theta}_i - \bar{\bm{\theta}}_t)^\top \nabla_{\bm{\theta}} L(\bar{\bm{\theta}}_t).
\end{align}
These assumptions imply $ \nabla_{\bm{\theta}} L(\bm{\theta}_i) \approx \nabla_{\bm{\theta}} L(\bar{\bm{\theta}}_t) $.  
By summing Eq.~\eqref{approx0} over $ i \in \{t\!-\!n\!+\!1, \ldots, t\} $, we obtain the following equation:
\begin{align}
& nL(\bar{\bm{\theta}}_t) \notag \\
\approx & \sum_i L(\bm{\theta}_i) + \left(n\bar{\bm{\theta}_t} - \sum_i \bm{\theta}_i\right)^\top \nabla_{\bm{\theta}} L(\bar{\bm{\theta}}_t) \notag \\
= & \sum_i L(\bm{\theta}_i),
\end{align}
since $ n\bar{\bm{\theta}}_t = \sum_i \bm{\theta}_i $.  
Thus, the evaluation score at $ \bar{\bm{\theta}}_t $ approximates the average score across the previous checkpoints.  
The expected value of the evaluation score under majority voting also corresponds to the average score across all models used in the ensemble.

Let $ L_t = L(\bm{\theta}_t) $ and $ \bar{L}_t = L(\bar{\bm{\theta}}_t) \approx \frac{1}{n} \sum_i L_i $.  
Note that
\begin{align}
\bar L_{t+1}-\bar L_t
\approx \frac{1}{n}\sum_{i=t-n+1}^{t}\bigl(L_{i+1}-L_{i}\bigr).   
\end{align}
By the triangle inequality,
\begin{align}
\bigl|\bar L_{t+1}-\bar L_t\bigr|
\;\le\;
\frac{1}{n}\sum_{i=t-n+1}^{t}\bigl|L_{i+1}-L_{i}\bigr|.    
\end{align}
Summing over all $t$ gives
\begin{align}
\mathrm{MTV}(\bar{\bm{\Theta}})
&=\frac{1}{m\!-\!1}\sum_{t=1}^{m-1}\bigl|\bar L_{t+1}-\bar L_t\bigr| \notag \\
&\le
\frac{1}{m\!-\!1}\sum_{t=1}^{m-1}\frac{1}{n}\sum_{i=t-n+1}^{t}\bigl|L_{i+1}-L_{i}\bigr| \notag \\
&=\frac{1}{n}\sum_{t=1}^{n}\frac{1}{m\!-\!1}\sum_{i=1}^{m-1}\bigl|L_{i+1}-L_{i}\bigr| \notag \\
&=\frac{1}{n}\sum_{t=1}^{n}\mathrm{MTV}(\bm{\Theta}) \notag \\
&=\mathrm{MTV}(\bm{\Theta}).
\end{align}

\begin{figure*}[t]
\captionsetup[subfigure]{justification=centering}
  \centering
  \begin{minipage}[b]{\subfigwidth}
    \centering
    \includegraphics[width=\textwidth]{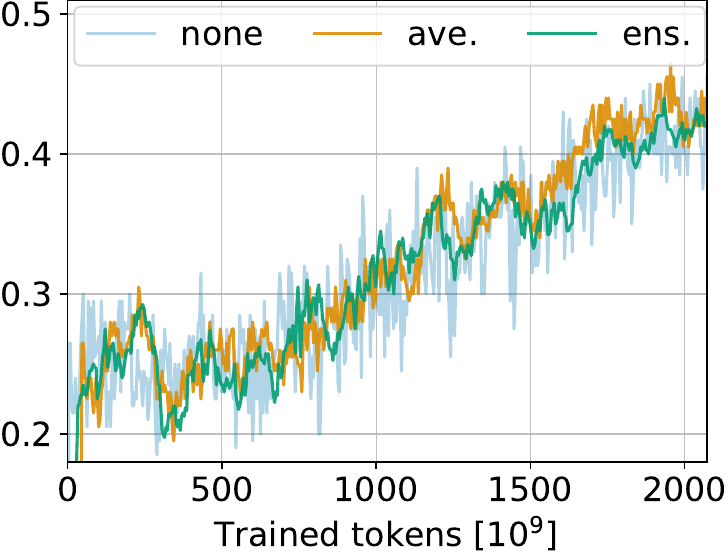}
    \subcaption{HE (Human Examination)}\label{fig:stabled_he}
  \end{minipage}
  \hfill
  \begin{minipage}[b]{\subfigwidth}
    \centering
    \includegraphics[width=\textwidth]{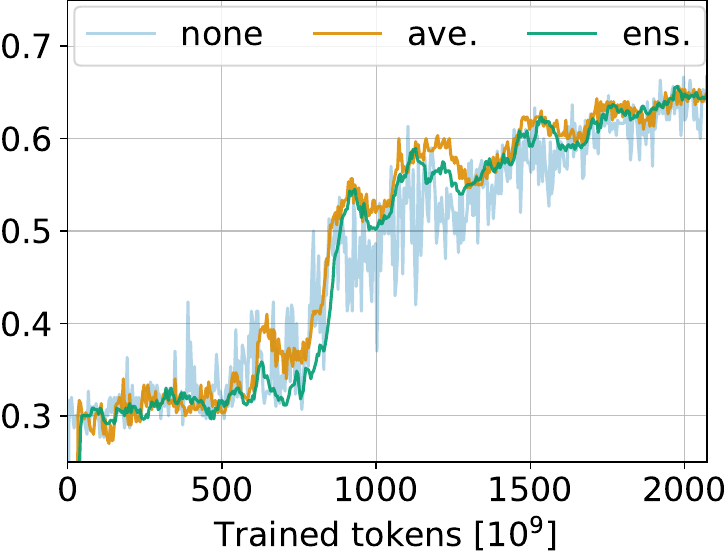}
    \subcaption{MC (Multiple Choice QA)}\label{fig:stabled_mc}
  \end{minipage}
  \hfill
  \begin{minipage}[b]{\subfigwidth}
    \centering
    \includegraphics[width=\textwidth]{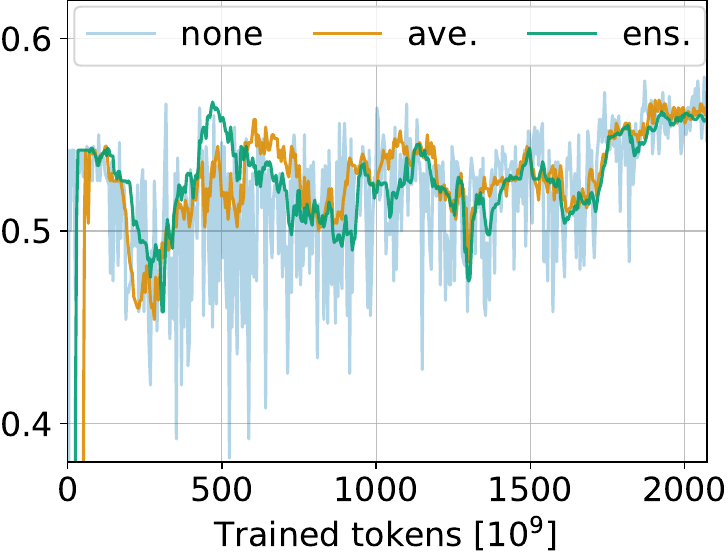}
    \subcaption{NLI (Natural Language Inference)}\label{fig:stabled_nli}
  \end{minipage}
  \caption{Progression of downstream task scores for the 13B model when applying stabilization methods. ``none'' represents no stabilization method, ``ave.'' represents checkpoint averaging, and ``ens.'' represents checkpoint ensemble scores. The window size for stabilization methods was set at 20.}
  \label{fig:stabled_score}
\end{figure*}

\subsection{Effect on Task-Level Scores}
We first evaluate the impact of stabilization methods on downstream task performance at the category level.
\cref{tab:stabilized_score} summarizes the mean and standard deviation of scores for the 13B model across all task categories, measured over the final 20\% of training checkpoints.
This final-phase evaluation focuses on the later training stages, which are typically used for model selection and benchmark reporting.

Applying checkpoint averaging leads to improvements in both mean scores and stability (i.e., lower standard deviation) in most task categories, including the overall score.
These effects become more pronounced with larger averaging window sizes.
A similar trend is observed with the checkpoint ensemble, which also reduces score variance and enhances mean accuracy.
Both results support our hypothesis that checkpoint integration suppresses short-term fluctuations and improves reliability in evaluation.
When comparing the two methods, checkpoint averaging tends to yield greater improvements in mean performance, particularly for larger windows.
These trends are consistent across other model sizes as well.

While \cref{tab:stabilized_score} highlights the aggregate improvements in performance and stability, \cref{fig:stabled_score} provides a qualitative view of how the stabilization methods affect score trajectories throughout training.  
The figure visualizes the 13B model's performance in three representative categories (HE, MC, and NLI) under different stabilization settings.  
With a window size of 20, both checkpoint averaging and ensemble methods visibly suppress score fluctuations across training steps, indicating enhanced stability throughout pretraining.

\subsection{Effect on Example-Level Scores}
\begin{figure*}[t]
\captionsetup[subfigure]{justification=centering}
  \centering
  \begin{minipage}[b]{1.0\linewidth}
    \centering
    \includegraphics[width=\textwidth]{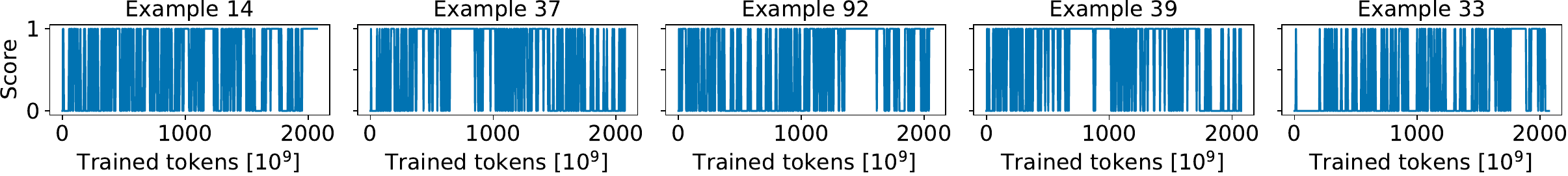}
    \subcaption{No stabilization method applied.}
    \label{fig:13b_jcommonsenseqa}
  \end{minipage}
  \\
  \vspace{0.8em}
  \begin{minipage}[b]{1.0\linewidth}
    \centering
    \includegraphics[width=\textwidth]{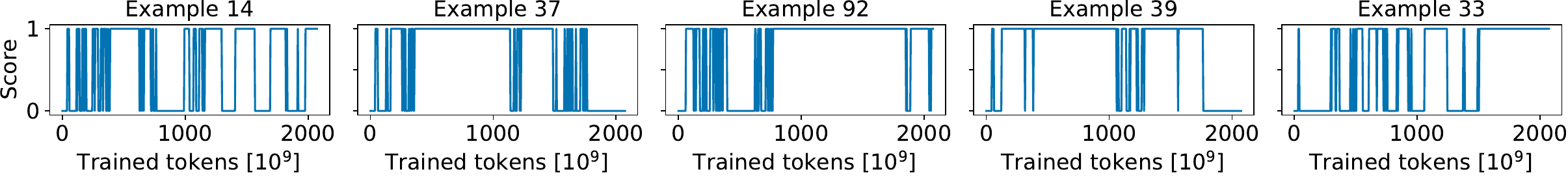}
    \subcaption{Checkpoint averaging with window size 20.}
    \label{fig:13b-rolling20_jcommonsenseqa}
  \end{minipage}
  \\
  \vspace{0.8em}
  \begin{minipage}[b]{1.0\linewidth}
    \centering
    \includegraphics[width=\textwidth]{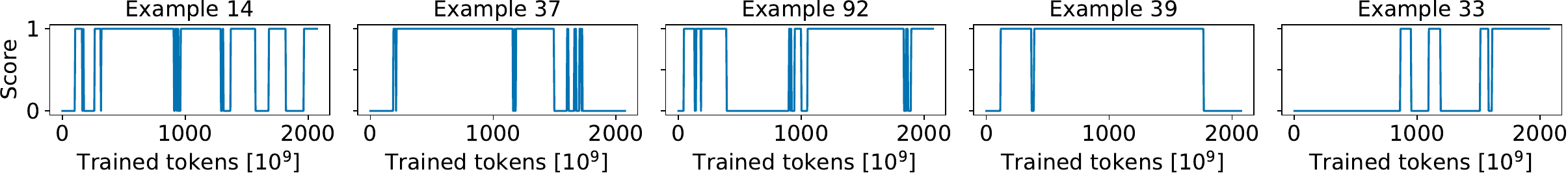}
    \subcaption{Checkpoint ensemble with window size 20.}
    \label{fig:13b-voting20_jcommonsenseqa}
  \end{minipage}

  \caption{Example-level score trajectories for JCommonsenseQA in the 13B model with and without stabilization methods. The examples shown were selected based on having the highest MTV under the no-stabilization setting.}
  \label{fig:example_level_stabilization}
\end{figure*}

We now examine whether the stabilization effects also hold at the example level.  
\cref{fig:example_level_stabilization} illustrates score trajectories for five representative examples from JCommonsenseQA (MC) before and after applying stabilization methods.  
The examples were selected based on having the highest MTV under the no-stabilization setting.  
As shown, both checkpoint averaging and ensemble yield qualitatively smoother predictions, indicating reduced instability across training.

\begin{figure*}[t]
\centering
\begin{minipage}[b]{0.48\linewidth}
  \centering
  \includegraphics[width=\linewidth]{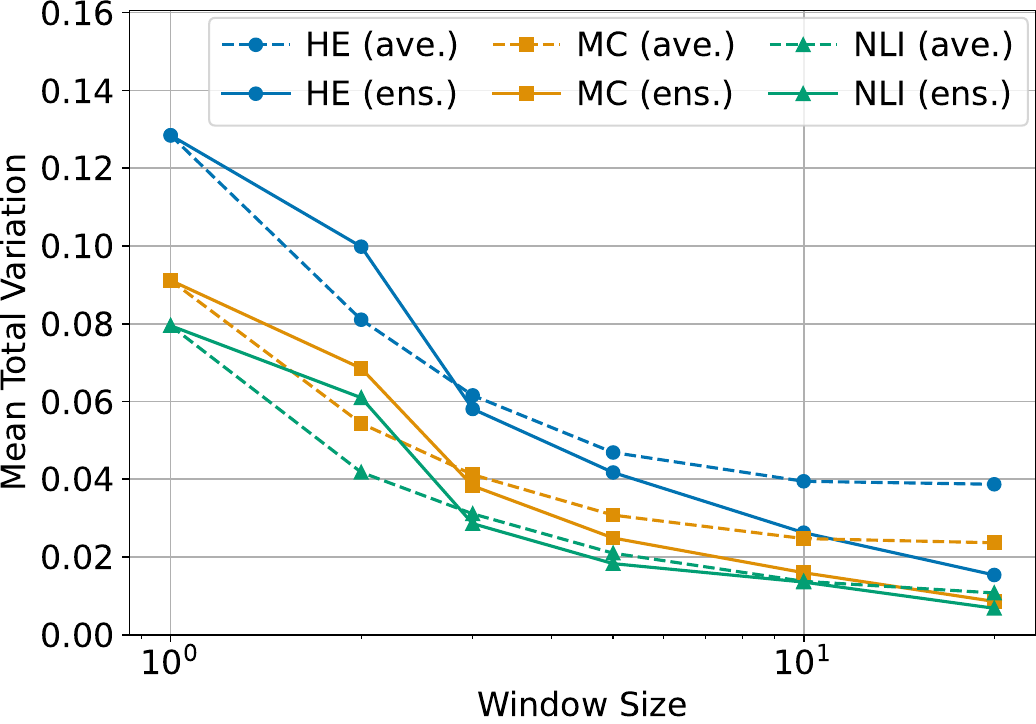}
  \subcaption{Mean total variation (MTV)}\label{fig:tv_by_window_size}
\end{minipage}
\hfill
\begin{minipage}[b]{0.48\linewidth}
  \centering
  \includegraphics[width=\linewidth]{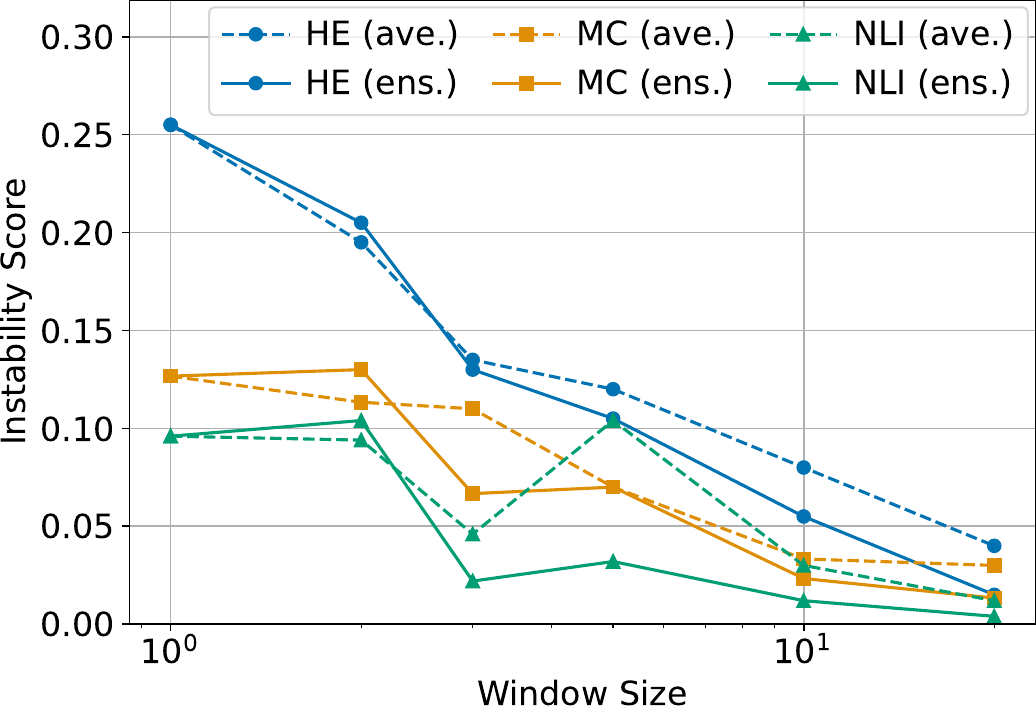}
  \subcaption{Instability score (IS)}\label{fig:instability_score_by_window_size}
\end{minipage}
\caption{Two metrics of downstream performance instability for the 13B model across task categories, measured over the final 20\% of training checkpoints.  
Both metrics decrease as the window size increases, showing that checkpoint averaging (``ave.'') and checkpoint ensemble (``ens.'') effectively mitigate instability.}
\label{fig:tv_by_window_size_on_multiple_choice_category}
\end{figure*}

To quantitatively evaluate this effect, we measure example-level instability using both the mean total variation (MTV) and the instability score (IS), following the same procedure described in \cref{sec:instability_quantification}.  
\cref{fig:tv_by_window_size_on_multiple_choice_category} shows how these two metrics change with different window sizes for the 13B model in three representative task categories: MC, NLI, and HE.\footnote{Results for the remaining task categories are presented in Appendices~\ref{appendix:mtv_all_categories} and \ref{appendix:instability_score_all_categories}.}  
A window size of 1 corresponds to the unstabilized baseline.  
As the window size increases, both metrics consistently decrease for all categories, confirming that checkpoint averaging and checkpoint ensemble effectively mitigate example-level instability.  
This trend is consistent across both reference-based (MTV) and reference-free (IS) metrics, highlighting that checkpoint integration reduces instability in model outputs beyond a specific evaluation metric.  
Overall, these results align with our task-level findings and demonstrate that the stabilizing effects of checkpoint integration extend to individual examples.

\section{Related Work}
\label{sec:related_work}
\paragraph{Stability in Language Model Training.}
Stability in language model training has primarily been investigated through the behavior of training loss.
Challenges such as loss divergence and large fluctuations, as well as inconsistency due to sensitivity to random initialization, have been widely recognized.
To address these issues, various architectural and optimization strategies have been proposed to improve convergence and reduce variance across training runs~\cite{chowdhery2023palm,rybakov2024methodsimprovingllmtraining,takase2024spikemorestabilizingpretraining}.
In addition, studies have examined token-level prediction probabilities during training and found that high-frequency tokens tend to be learned more reliably~\cite{chang-etal-2024-characterizing}.
However, these studies have mainly focused on internal metrics and do not directly address fluctuations in downstream task performance throughout pretraining.

\paragraph{Progression of Downstream Task Performance in LLMs.}
Several recent studies have tracked how language models acquire capabilities during pretraining by analyzing downstream task performance over time~\cite{biderman2023pythia, xia-etal-2023-training, yang2024finelinenavigatinglarge, gadre2024languagemodelsscalereliably}.
For example, \citet{biderman2023pythia} examined the learning curves of models of different sizes in the Pythia suite, while \citet{yang2024finelinenavigatinglarge} analyzed example-level learning trajectories to uncover when and how individual examples are learned.
\citet{gadre2024languagemodelsscalereliably} showed that larger models tend to exhibit smoother progress on evaluation tasks. 
These works focus on identifying learning phases or scaling behaviors.
In contrast, our study concentrates on the short-term instability of downstream task scores, such as frequent reversals in performance, which can occur even in the later stages of pretraining.

\paragraph{Inference Using Multiple Checkpoints.}
Using multiple checkpoints from a single training run has been proposed as a way to improve robustness and stability.
Two common techniques are checkpoint averaging, which averages model weights, and checkpoint ensemble, which aggregates predictions from multiple checkpoints~\cite{junczys-dowmunt-etal-2016-neural, chen2017checkpointensemblesensemblemethods, Liu2018ACS}.
These methods are primarily used to improve the final performance of natural language processing models.
Notably, \citet{gao-etal-2022-revisiting} reported that applying checkpoint averaging to neural machine translation models also led to improved stability in translation quality.
However, the effectiveness of such checkpoint-based integration methods for checkpoints during the pretraining of large language models remains largely unexplored.
Our work addresses this gap by empirically demonstrating that simple checkpoint integration strategies can effectively mitigate downstream performance instability during LLM pretraining, without requiring any changes to the training procedure.

\section{Conclusion}
This study analyzed the instability of downstream task performance during the pretraining of large language models, focusing on performance fluctuations across checkpoints.  
We found that such instability occurs widely across tasks and model sizes.  
These findings suggest that downstream performance can fluctuate significantly even in the later stages of training, which can undermine the reliability and reproducibility of evaluation results.

To address this, we investigated post-processing approaches that integrate multiple checkpoints.  
Both checkpoint averaging and ensemble methods effectively reduced performance fluctuations at both the task and example levels, without modifying the training procedure.  
While checkpoint averaging contributed more to improving average scores, checkpoint ensemble was more effective in stabilizing example-level predictions.
Given its lower inference cost, checkpoint averaging is particularly well-suited for practical deployment.

\section*{Limitations}
Our work demonstrates that instability in downstream task performance is prevalent during LLM pretraining, and that simple checkpoint integration methods can effectively mitigate this issue at both the task and example levels, without requiring changes to the training procedure.
However, to further improve robustness and generality, there remain several limitations to be addressed.  
First, the integration methods explored in this work do not incorporate model confidence or prediction uncertainty, which could offer additional gains in stabilization.  
Second, our analysis focuses on models trained on a single multilingual corpus and does not extend to instruction-tuned or alignment-based models.

Expanding the scope of checkpoint integration to include adaptive strategies, such as confidence-weighted averaging, and evaluating models across diverse pretraining regimes would be valuable next steps.  
Additionally, analyzing how output instability relates to prediction probability dynamics and internal representations could offer deeper insights into the nature of learning fluctuations.  
Understanding whether the observed instability persists or transforms in downstream-tuned models also remains an important open question.

\section*{Acknowledgments}

This work was supported by the “R\&D Hub Aimed at Ensuring Transparency and Reliability of Generative AI Models” project of the Ministry of Education, Culture, Sports, Science and Technology.

\bibliography{custom}

\appendix

\section{Detailed Settings}
\label{appendix:detailed_settings}
\cref{tab:checkpoints} summarizes the number of parameters, batch sizes, maximum sequence lengths, and the training steps at which checkpoints were obtained for each model.
The checkpoints were saved at regular intervals ranging from several hundred to approximately a thousand steps.

\cref{tab:llm-jp-eval_dataset} lists all downstream task categories, datasets, and the evaluation metrics used.
Each task category typically comprises multiple datasets.
For example, the MC (multiple choice question answering) category includes JCommonsenseQA, JCommonsenseMorality, and KUCI.

\begin{table*}[t]
\centering
\small
\begin{tabular}{@{}rlrr@{}}
\toprule
Model size & Checkpointed steps (total) & Batch size & Max tokens \\
\midrule
150M & 0, 1, $\cdots$, 9, 10, $\cdots$, 90, 100, $\cdots$, 900, 1000, $\cdots$, 988000, 988240 (1,017 total) & 512 & 4,096 \\
440M & 0, 1, $\cdots$, 9, 10, $\cdots$, 90, 100, $\cdots$, 900, 1000, $\cdots$, 988000, 988240 (1,017 total) & 512 & 4,096 \\
980M & 0, 1, $\cdots$, 9, 10, $\cdots$, 90, 100, $\cdots$, 900, 1000, $\cdots$, 988000, 988240 (1,017 total) & 512 & 4,096 \\
1.8B & 0, 1, $\cdots$, 9, 10, $\cdots$, 90, 100, $\cdots$, 900, 1000, $\cdots$, 988000, 988240 (1,017 total) & 512 & 4,096 \\
3.7B & 0, 1, $\cdots$, 9, 10, $\cdots$, 90, 100, $\cdots$, 900, 1000, $\cdots$, 494000, 494120 (523 total) & 1,024 & 4,096 \\
7.3B & 0, 1, $\cdots$, 9, 10, $\cdots$, 90, 100, $\cdots$, 900, 1000, $\cdots$, 494000, 494120 (523 total) & 1,024 & 4,096 \\
13B  & 0, 1, $\cdots$, 9, 10, $\cdots$, 90, 100, $\cdots$, 900, 1000, $\cdots$, 494000, 494120 (523 total) & 1,024 & 4,096 \\
\bottomrule
\end{tabular}
\caption{Training configurations and saved checkpoints for each model size.}
\label{tab:checkpoints}
\end{table*}

\begin{table*}[t]
\small
\centering
\begin{tabular}{@{}llll@{}}
\toprule
Task category & Dataset & Sub task & Evaluation metric \\
\midrule
NLI (natural language inference) & \hyperref[dataset:jamp]{Jamp} & - & Exact match \\
& \hyperref[dataset:janli]{JaNLI} & - & Exact match \\
& \hyperref[dataset:jnli]{JNLI} & - & Exact match \\
& \hyperref[dataset:jsem]{JSeM} & - & Exact match \\
& \hyperref[dataset:jsick]{JSICK} & - & Exact match \\
\midrule
QA (question answering) & \hyperref[dataset:jemhopqa]{JEMHopQA} & - & Char. F1 \\
& \hyperref[dataset:niilc]{NIILC} & - & Char. F1 \\
\midrule
RC (reading comprehension) & \hyperref[dataset:jsquad]{JSQuAD} & - & Char. F1 \\
\midrule
MC (multiple choice question answering) & \hyperref[dataset:jcm]{JCommonsenseMorality} & - & Exact match \\
& \hyperref[dataset:jcommonsenseqa]{JCommonsenseQA} & - & Exact match \\
& \hyperref[dataset:kuci]{KUCI} & - & Exact match \\
\midrule
EL (entity linking) & \hyperref[dataset:chabsa]{chABSA} & - & Set F1 \\
\midrule
FA (fundamental analysis) & \hyperref[dataset:wikipedia]{Wikipedia Annotated Corpus} & NER & Set F1 \\
& & PAS & Set F1 \\
& & Coreference & Set F1 \\
& & Dependency & Set F1 \\
& & Reading & Char. F1 \\
\midrule
MR (mathematical reasoning) & \hyperref[dataset:mawps]{MAWPS} & - & Exact match \\
\midrule
MT (machine translation) & \hyperref[dataset:alt]{ALT} & Ja$\to$En & COMET \cite{rei-etal-2020-comet} \\
& & En$\to$Ja & COMET \\
& \hyperref[dataset:wikicorpus]{WikiCorpus} & Ja$\to$En & COMET \\
& & En$\to$Ja & COMET \\
\midrule
HE (human examination) & \hyperref[dataset:mmlu]{MMLU} & - & Exact match \\
& \hyperref[dataset:jmmlu]{JMMLU} & - & Exact match \\
\bottomrule
\end{tabular}
\caption{Task categories, datasets, subtasks, and evaluation metrics used in the llm-jp-eval framework. Each category may include multiple datasets, each evaluated with a task-specific metric.}
\label{tab:llm-jp-eval_dataset}
\end{table*}

\section{Additional Results}
\label{appendix:additional_results}

\subsection{Mean Total Variation for All Task Categories}
\label{appendix:mtv_all_categories}
To supplement the main results shown in \cref{fig:tv_by_window_size_on_multiple_choice_category}, we report the mean total variation (MTV) across different window sizes for all task categories using the 13B model.  
As described in \cref{sec:instability_quantification}, MTV quantifies the average fluctuation in example-level scores across checkpoints, with lower values indicating greater temporal stability in predictions.

\begin{figure}[t]
\centering
\includegraphics[width=1.0\linewidth]{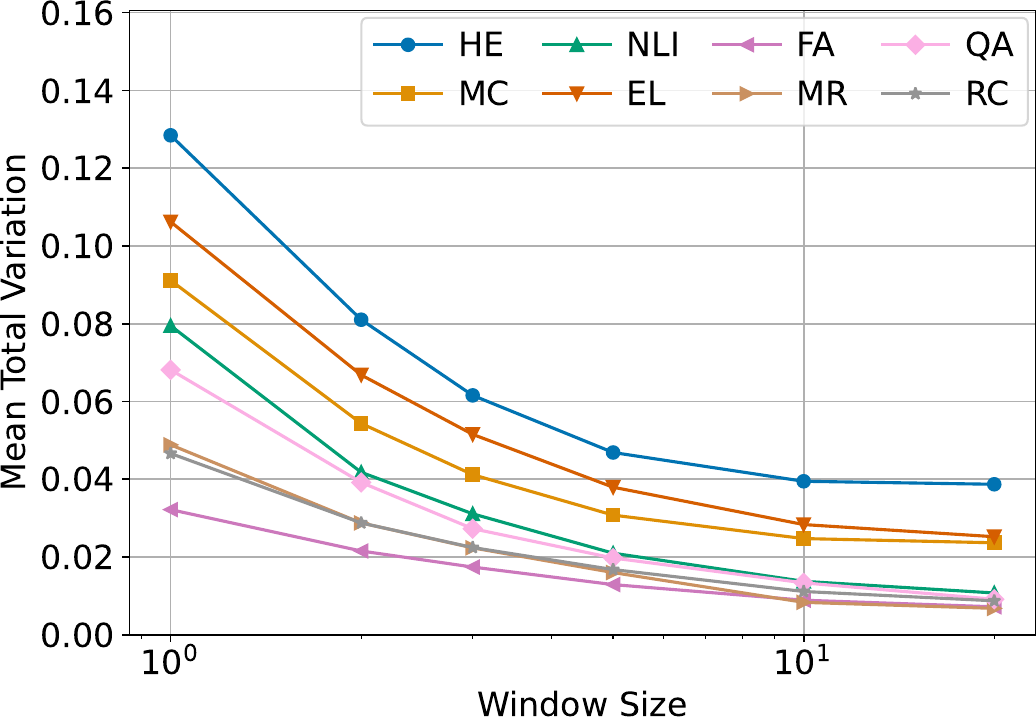}
\caption{Mean total variation for the 13B model across all task categories using checkpoint averaging. This figure complements the main text by providing results for the remaining categories and enabling direct comparison with HE, MC, and NLI.}
\label{fig:tv_by_window_size_on_rolling}
\end{figure}

While \cref{fig:tv_by_window_size_on_multiple_choice_category} focused on three representative categories (HE, MC, and NLI), \cref{fig:tv_by_window_size_on_rolling} presents a complementary overview that includes the remaining six categories alongside the previously shown ones for comparison.  
Since the results of checkpoint ensemble have already been presented for all categories, this figure exclusively reports the results of checkpoint averaging.

The trends observed here are consistent with the findings in the main text: increasing the averaging window size reduces MTV across categories, indicating the stabilizing effect of checkpoint averaging during LLM pretraining.

\subsection{Instability Score Across Tasks}
\label{appendix:instability_score_all_categories}
In addition to the MTV results presented in the main text, we report complementary results using the instability score (IS) introduced in \cref{sec:instability_quantification}.  
Unlike MTV, which measures reference-based score fluctuations, IS directly evaluates the variability of model outputs by computing dissimilarity between consecutive predictions.

\begin{figure}[t]
\centering
\includegraphics[width=1.0\linewidth]{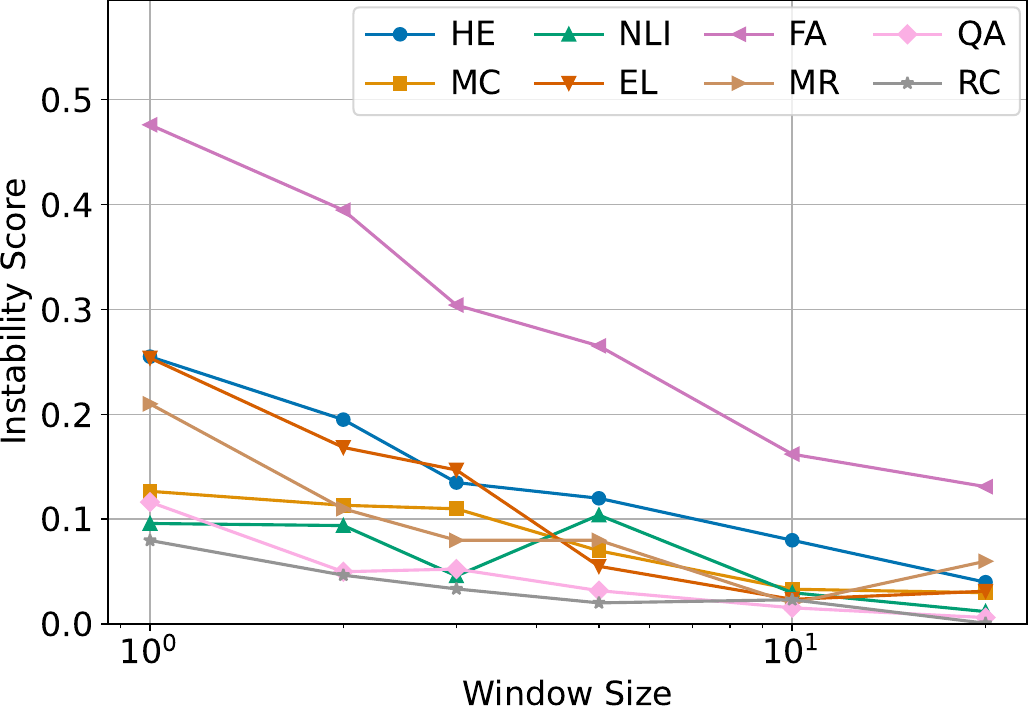}
\caption{Instability score for the 13B model across different task categories using checkpoint averaging. Larger window sizes consistently reduce output instability.}
\label{fig:instability_score_on_rolling_by_window_size}
\end{figure}

\cref{fig:instability_score_on_rolling_by_window_size} illustrates the effect of checkpoint averaging across different window sizes.  
We observe a consistent reduction in IS as the window size increases, confirming that averaging helps smooth output transitions across checkpoints.

\section{Used Data and Software}
\subsection{Data}
\paragraph{Jamp} \phantomsection\label{dataset:jamp}
created by \citet{sugimoto-etal-2023-jamp}. License: CC BY-SA 4.0. \url{https://github.com/tomo-ut/temporalNLI_dataset}

\paragraph{JaNLI} \phantomsection\label{dataset:janli}
created by \citet{yanaka-EtAl:2021:blackbox}. License: CC BY-SA 4.0. \url{https://github.com/verypluming/JaNLI}

\paragraph{JNLI} \phantomsection\label{dataset:jnli}
created by \citet{kurihara-etal-2022-jglue}. License: CC BY-SA 4.0. \url{https://github.com/yahoojapan/JGLUE}

\paragraph{JSeM} \phantomsection\label{dataset:jsem}
created by \citet{kawazoe2015entailment}. License: Available for research use. \url{https://github.com/DaisukeBekki/JSeM}

\paragraph{JSICK} \phantomsection\label{dataset:jsick}
created by \citet{yanaka-mineshima-2022-compositional}. License: CC BY 4.0. \url{https://github.com/verypluming/JSICK}

\paragraph{JEMHopQA} \phantomsection\label{dataset:jemhopqa}
created by \citet{ishii-etal-2024-jemhopqa}. License: CC BY-SA 4.0. \url{https://github.com/aiishii/JEMHopQA}

\paragraph{NIILC} \phantomsection\label{dataset:niilc}
created by Sekine et al. License: CC BY-SA 4.0. \url{https://mynlp.is.s.u-tokyo.ac.jp/niilc-qa/}

\paragraph{JSQuAD} \phantomsection\label{dataset:jsquad}
created by \citet{kurihara-etal-2022-jglue}. License: CC BY-SA 4.0. \url{https://github.com/yahoojapan/JGLUE}

\paragraph{JCommonsenseMorality} \phantomsection\label{dataset:jcm}
created by Takeshita et al. License: MIT. \url{https://github.com/Language-Media-Lab/commonsense-moral-ja}

\paragraph{JCommonsenseQA} \phantomsection\label{dataset:jcommonsenseqa}
created by \citet{kurihara-etal-2022-jglue}. License: CC BY-SA 4.0. \url{https://github.com/yahoojapan/JGLUE}

\paragraph{KUCI} \phantomsection\label{dataset:kuci}
created by \citet{omura-etal-2020-method}. License: CC BY-SA 4.0. \url{https://github.com/ku-nlp/KUCI}

\paragraph{chABSA} \phantomsection\label{dataset:chabsa}
created by Kubo et al. License: CC BY 4.0. \url{https://github.com/chakki-works/chABSA-dataset}

\paragraph{Wikipedia Annotated Corpus} \phantomsection\label{dataset:wikipedia}
created by Kawahara et al. License: CC BY-SA 4.0. \url{https://github.com/ku-nlp/WikipediaAnnotatedCorpus}

\paragraph{MAWPS} \phantomsection\label{dataset:mawps}
created by \citet{horio2023cot}. License: Apache-2.0 license. \url{https://github.com/nlp-waseda/chain-of-thought-ja-dataset}

\paragraph{Asian Language Treebank (ALT) Parallel Corpus} \phantomsection\label{dataset:alt}
created by \citet{thu-etal-2016-introducing}. License: CC BY-NC-SA 4.0. \url{https://www2.nict.go.jp/astrec-att/member/mutiyama/ALT/}

\paragraph{WikiCorpus} \phantomsection\label{dataset:wikicorpus}
created by NICT. License: CC BY-SA 3.0. \url{https://alaginrc.nict.go.jp/WikiCorpus/}

\paragraph{MMLU} \phantomsection\label{dataset:mmlu}
created by \citet{hendryckstest2021}. License: MIT. \url{https://github.com/hendrycks/test}

\paragraph{JMMLU} \phantomsection\label{dataset:jmmlu}
created by Kawazoe et al. License: CC BY-SA 4.0. \url{https://github.com/nlp-waseda/JMMLU}

\paragraph{LLM-jp Corpus v3} created by LLM-jp. License: Available for research use. \url{https://gitlab.llm-jp.nii.ac.jp/datasets/llm-jp-corpus-v3/}

\subsection{Software}
\paragraph{llm-jp-eval} created by Han et al. License: Apache-2.0 license. \url{https://github.com/llm-jp/llm-jp-eval}

\end{document}